\documentclass[a4paper]{article}

\usepackage[english]{babel}
\usepackage[utf8x]{inputenc}
\usepackage[T1]{fontenc}

\usepackage[a4paper,top=3cm,bottom=2cm,left=3cm,right=3cm,marginparwidth=1.75cm]{geometry}

\usepackage{amsmath}
\usepackage{graphicx}
\usepackage[colorinlistoftodos]{todonotes}

\usepackage{authblk}
\title{Nonlinear control of a swinging pendulum on a wheeled mobile robot with nonholonomic constraints}
\author{Nikhil Potu Surya Prakash}
\affil{Department of Mechanical Engineering \\
	   University of Michigan,\\
       Ann Arbor, MI-48105,\\
       Email:nikhilps@umich.edu
}

\date{}                     
\begin{document}
\maketitle

\begin{abstract}
In this paper, we propose a nonlinear control strategy for swinging up a pendulum to its upright equilibrium position by shaping its swinging energy along with regulating the cart to a desired location. While the base of a usual cart-pole system is restricted to move in a straight line, the present system is allowed to move in the x-y plane with a nonholonomic consraint that its allowable velocity is only along its orientation. A simple time invariant control law has been presented and its effectiveness has been demonstrated using numerical experiments.
\end{abstract}

\section{Introduction}
Most of the controllers that are derived for nonholonomic systems are either time dependent or at kinematic level. But using kinematics alone, the dynamic structure and properties of the system are often left out.In a case like the present system where the dynamics of the pendulum are coupled with the dynamics of the wheeled mobile robot upon which it is appended, kinematic level models and control strategies prove to be futile and there arises a need to depend on the dynamics of the complete system.In an attempt to exploit the rich dynamics of underactuated systems with nonholonomic constraints, such a system has been modelled.

Energy shaping is a technique that has been proven to be very effective and has been extensively studied through implementations on various underactuated systems with pendulums. In energy shaping, first a desired energy of the whole system or a part of the system is formulated to which the system's energy or a part of it is converged. This ensures bringing the states of the system close to the desire states from almost all initial conditions. Once the system is put in the orbit (of that particular energy level), a linear controller is implemented once the states of the system come close to the desired states. In the pendulum's energy shaping control [4], the pendulum's energy at the upright equilibrium position is considered and the energy of the whole system is converged to this energy level, which basically is same as bringing the states of the system onto a homoclinic orbit which starts and ends at the same unstable equilibrium point.A lyapunov function can be constructed by taking the square of the error in the desired energy and by deriving a control law based on the error in the energy, it can be shown that the error asymptotically converges to zero. It must be noted that just bringing the states of the pendulum to the homoclinic orbit does not stabilize the system about its upright equilibrium point, since any small disturbance would make the control law to converge the system back to the homoclinic orbit which might require the pendulum to complete one full revolution and then come back to the upright equilibrium point. Therefore, a need to use a linear controller near the unstable equilibrium point arises. 

Other energy shaping controllers include swinging up an Acrobot to its upright equilibrium position [6], where first a partial feedback linearization of the system is conducted and then an energy shaping controller is implemented. Another controller for an Acrobot [5] uses a slightly modified lyapunov function which along with the square of energy includes a term quadratic in the angular position and angular velocity of the bottom link. The idea there was to make the Acrobot behave like a single pendulum and swing it to the upright position. Although, our present system is closer to a cart-pole [7] where the pendulum's energy alone is considered and a pd controller is added to regulate the position of the cart to a desired position. Other notable systems include a Pendubot [8], Furuta pendulum [9,10], inertia wheel pendulum [11] etc.   
\begin{figure}[h!]
	\centering
    \includegraphics[width=5in]{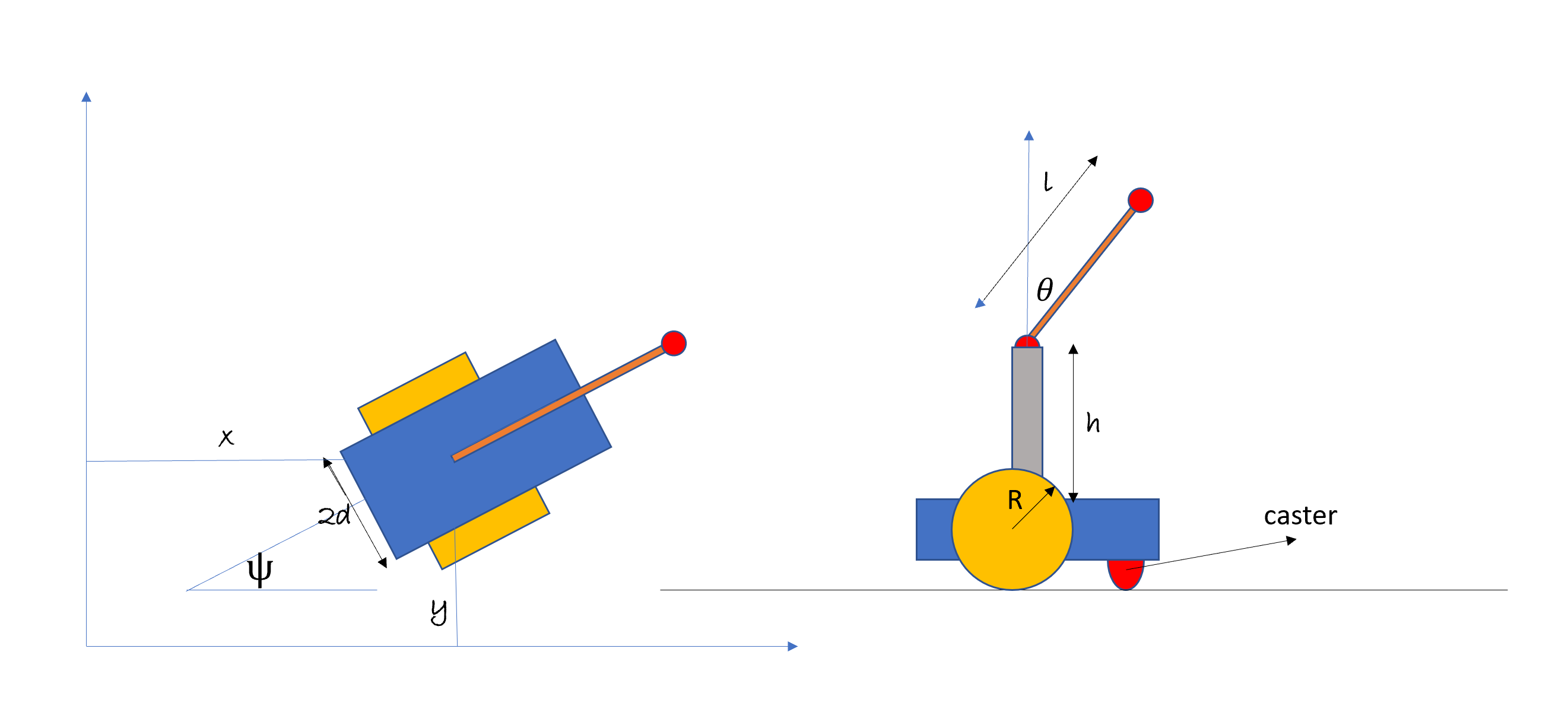}
\caption{\label{fig:Motion} Schematic of a pendulum on a wheeled mobile robot}
	\label{schematic1}
\end{figure}

\section{Model Description of the system}
The system in consideration (see fig [1]) has a wheeled mobile robot (WMR) carrying a pendulum mounted on top which is allowed to swing in the plane formed by the unit vector normal to the plane in which the WMR moves and the unit vector along the direction of motion of the WMR. the WMR is a differntial drive robot with two wheels to which torques can be applied to accelarate and rotate the robot. A smooth caster is introduced to make the WMR a stable system. The assumptions on the system are that the wheels are only allowed to roll without slipping and the system cannot move in a direction perpendicular to its orientation which form the nonholonomic constraints on the system.\newline

The degrees of freedom of the system are the cartesian coordinates of the positions of the cart (x and y), the orientation or the yaw angle ($\psi$) of the WMR (with positive X axis of the inertial frame), the orientations of the wheels ($\phi_l$ - left wheel and $\phi_r$ - right wheel) and the angle made by the pendulum with the vertical ($\theta$). Therefore the system evolves on a manifold whose configuration space is $SE(2) \times S^1 \times S^1 \times S^1$ the state vector of the system is $[x,y,\psi,\phi_r,\phi_l,\theta,\dot x,\dot y,\dot \psi,\dot \phi_r, \dot \phi-l, \dot \theta]$. We assume that the motion of the skate is restricted to a plane with the same gravitational potential throughout and hence there will not be any change in the potential energy of the WMR as the it moves.

The nonholonomic constraints on the motion are given by the conditions that all the velocity of the WMR is along the direction in which it is oriented (no lateral slippage) and both the wheels roll without slipping.We now show how one can arrive at the nonholonomic constraints and derive the equations of motion of the system.
The position vectors of different parts of the system expressed in the inertial frame are given by
\begin{align}
r_{center \thinspace of \thinspace mass} =& (x,y,-h-l) \\
r_{right \thinspace wheel} =& (x+d sin\psi,y+d cos\psi,-h-l) \\
r_{left \thinspace wheel} =& (x-d sin\psi,y-d cos\psi,-h-l) \\
r_{pendulum \thinspace bob} =& (x+l sin\theta cos\psi,y+l sin\theta sin\psi,l-l cos\theta) 
\end{align}

Here h always remains constant and does not appear in the equations of motion. l is the length of the pendulum. Also since the potential energy of the WMR always remains constant, this can be omitted from the formulation.
Differentiating the position vectors, we obtain the velocities of the parts which are also expressed in the inertial frame.
\begin{align}
v_{center \thinspace of \thinspace mass} =& (\dot x,\dot y,0)\\
v_{right \thinspace wheel} =& (\dot x+d \dot \psi cos\psi,y-d \dot \psi sin\psi,0)\\
v_{left \thinspace wheel} =& (\dot x-d \dot \psi cos\psi,y+d \dot \psi sin\psi,0)\\
v_{pendulum \thinspace bob} =& (x+l \dot \theta cos\theta cos\psi-l \dot \psi sin \theta sin \psi,\nonumber \\
&\dot y+l \dot \theta cos \theta sin \psi+l \dot \psi sin\theta cos\psi,l \dot \theta sin\theta)
\end{align}

Therefore, the constraints can now be given by the following equations
\begin{align}
&-\dot x sin\psi+ \dot y cos\psi = 0 \\
&\dot x cos\psi + \dot y sin\psi + d\dot \psi - R\dot \phi_r = 0 \\
&\dot x cos\psi + \dot y sin\psi - d\dot \psi - R\dot \phi_l = 0
\end{align}

where 2d is the distance between the wheels assuming that the WMR is symmetric and R is the radius of the wheels.
With the positions,velocities and the angular velocities set up, its quite straight forward to find the kinetic and potential energies of the system and thereby the Lagrangian of the system.
The total kinetic energy (T) of the system can therefore be expressed as
\begin{align}
T =& \frac{1}{2} J_{w1} \dot \phi_r^2 + \frac{1}{2} J_{w2} \dot \psi^2 + \frac{1}{2}m_w((\dot x +d \dot \psi cos \psi)^2+(\dot y - d \dot \psi sin \psi)^2)+\frac{1}{2}m_w((\dot x -d \dot \psi cos \psi)^2+(\dot y + d \dot \psi sin \psi)^2)+ \nonumber\\ &\frac{1}{2}m_c(\dot x^2 + \dot y^2) + \frac{1}{2} J_c \dot \psi^2 + \frac{1}{2}m_p ((l \dot \theta cos \theta - l \dot \psi sin \theta sin \psi)^2+(l \dot \theta cos \theta sin \psi + l \dot \psi sin \theta cos \psi)^2+(-l \dot \theta sin \theta)^2)
\end{align}
and the potential energy(V) of the system by taking the datum at the pendulum bob when it is vertical is given by
\begin{align}
V = -m_pg(l-l cos \theta) - (m_c + 2m_w)g(l+h)
\end{align}

where $m_p$ is the mass of the pendulum bob, $m_c$ is the mass of the WMR without the wheels, $m_w$ is the mas of each wheel, $J_{w1}$ and $J_{w2}$ are the mass moments of inertia of each wheel about the major and minor axes respectively and $J_c$ is the mass moment of inertia of the WMR without the wheels about the vertical axis passing through its center of gravity. 
Hence the lagrangian (L) is given by $L = T-V$ 
The present system is subject to three nonholonomic constraints which can be written in the form
\begin{align}
A^{*}(q) \dot q = 0
\end{align}
Here $A^{*}(q)$ is a 3×6 matrix and q is a 6×1  column vector. At any configuration q, the set of all possible virtual displacements is defined to be the subspace of the tangent space to the configuration manifold at q consisting of vectors $\delta q$ that satisfy the constraints, i.e., the subspace $D_q$ defined by
$$ D_q = \left\{ \delta q \in T_qQ \thinspace | \thinspace A^{*}(q).\delta q = 0 \right \} $$

The system is controlled by two inputs i.e., torques on each wheel ( $\tau_l$ on the left wheel and $\tau_r$ on the right wheel).
Choosing $ q = (r,s) $ where r and s are local coordinates and the forces enter only through the r coordinates such that equation (14) can be rewritten as $ \delta s^a+A_{\alpha}^a \delta r = 0$
The Lagrange–d’Alembert equations of motion for the system are those determined by
$\delta \int_{a}^{b}L(q^i,\dot q^i) dt = 0$
where we choose variations $\delta q(t)$ of the curve q(t) that satisfy $ \delta q(t)  \in D_q $
for each t, $a ≤ t ≤ b$ , and $\delta q(a) = \delta q(b) = 0$.
This principle is supplemented by the condition that the curve q(t) itself
satisfy the constraints.
we take the variation $\delta q$ before
imposing the constraints; that is, we do not impose the constraints on the family of curves defining the variation. The usual arguments in the calculus
of variations show that this constrained variational principle is equivalent
to the equations
\begin{align}
\int_{a}^{b} (\frac{d}{dt} \frac{\partial L}{\partial \dot q}- \frac{\partial L}{\partial q}-f) \delta q dt = 0
\end{align}
substituting the variations in local coordinates we obtain the equations of motion to be 
$$ \frac{d}{dt} \frac{\partial L}{\partial \dot r}- \frac{\partial L}{\partial r}-f = A(r,s) (\frac{d}{dt} \frac{\partial L}{\partial \dot s}- \frac{\partial L}{\partial s}) $$
where $f = (\tau_l,\tau_r)$ 

Using the nonholonomic constraints (9-11) to eliminate $\dot \phi_r$ and $\dot \phi_l$, the Lagrangian can be reduced and the resulting system is found analogous to a pendulum on a knife edge whose configurati0on manifold is $ SE(2) \times S^1$
The equations of motion for the reduced system can be established as follows 
\begin{align}
    &(M+m) \ddot x + m l\ddot \theta cos \theta cos \psi -m l^2 \dot \theta^2 sin \theta cos \psi - m l^2 \dot \theta \dot \psi cos \theta sin \psi + \lambda sin \psi = F cos \psi \\
    &(M+m) \ddot y + m l\ddot \theta cos \theta sin \psi -m l^2 \dot \theta^2 sin \theta sin \psi + m l^2 \dot \theta \dot \psi cos \theta cos \psi - \lambda cos \psi = F sin \psi \\
    &ml^2 \ddot \theta + ml cos \theta (\ddot x cos \psi + \ddot y sin \psi) - ml^2 \dot \psi^2 sin\theta cos \theta - m g l sin\theta = 0 \\
    &(J+ml^2 sin^2 \theta) \ddot \psi +ml^2 \dot \theta \dot \psi sin 2\theta = \tau
\end{align}

where M is the total mass of the WMR, J is the effective mass moment of inertia of the 
Here $\lambda = M[2\dot \psi (\dot x cos \psi + \dot y sin \psi)+l\dot \theta \dot \psi cos \theta]$ is the lateral frictional force acting on the WMR which can easily be computed using Newton's laws. $F = (\tau_l + \tau_r)/2d$ is the effective force on the WMR and $\tau = \tau_r-\tau_l$ is the effective moment on the system.

\section{Control design}
In this section we present a control strategy to bring the pendulum to its upright position along with regulating the WMR to a desired position. Without loss of generality we assume the desired position to be the origin in our further discussions. 
The swing up energy of the pendulum is defined as follows
\begin{align}
E = \frac{1}{2}ml^2 \dot \theta^2 - mgl(1-cos\theta)
\end{align}
It can easily be seen that the swing up energy when the pendulum is stationary with respect to the WMR at its upright position $ E = 0$. Once the states of the pendulum are such that they are on this swing up energy level $E = 0$, the pendulum reaches its upright position with with zero velocity in a finite time.
taking the derivative of the swing up energy we obtain
\begin{align}
    \dot E &= ml^2\dot \theta \ddot \theta - mgl sin \theta \dot \theta \nonumber \\
    &= (ml^2\ddot \theta - mgl sin \theta ) \dot \theta
\end{align}
$(16) \times cos\psi+(17) \times sin\psi$ gives us
\begin{align}
(M+m)(\ddot x cos \psi + \ddot y sin \psi) + ml \ddot \theta cos \theta - ml\dot \theta^2 sin \theta = F
\end{align}
Substituting (18) and (22) in (20) gives us 
\begin{align}
\dot E &= [mlcos \theta (\ddot x cos \psi + \ddot y sin \psi ) - ml^2 \dot \psi^2 sin \theta cos \theta] \dot \theta \nonumber \\
&=ml[(\ddot x cos \psi + \ddot y sin \psi ) -  l\dot \psi^2 sin \theta]cos \theta \dot \theta 
\end{align}
It is easy to see that $\ddot x cos \psi + \ddot y sin \psi$ is the acceleration of the WMR along the direction of its orientation ($a_l$). F can be applied in such a way that 
\begin{align}
    a_l = l \dot \psi^2 sin \theta cos \theta  - k_E \dot \theta cos \theta  E
\end{align}
for any positive value $k_E$.
Such a value ensures that the swing up energy exponentially converges to the orbit $ E = 0$ for almost all initial conditions except the ones starting at the downright position with zero velocity with respect to the WMR, since now
\begin{align}
    \dot E = -mlk_E \dot \theta^2 cos^2 \theta 
\end{align}
Constructing a lyapunov function in the following way proves that the swing up energy converges to zero. 
\begin{align}
    V_E = \frac{1}{2} E^2
\end{align}
The lyapunov function here is a simple quadratic function of the energy which is a positive definite function.The time rate of change of the lyapunov function is given by
\begin{align}
    \dot V_E &= E \dot E \nonumber \\
    &= -mlk_EE^2 \dot \theta^2 cos^2 \theta
\end{align}
Let D be the set of feasible initial conditions for the above control law to work.Then D can be defined as

\begin{align}
    D = \left \{(\theta,\dot \theta) \thinspace | \thinspace  \dot \theta \neq 0 \thinspace and \thinspace  \theta \neq (2n+1) \pi, n \in Z \right\}
\end{align}
Define $S \subseteq D $ such that
\begin{align}
    S &= \left \{(\theta,\dot \theta) \thinspace | \thinspace \dot V_E = 0 \right\} \nonumber \\
     &= \left \{(\theta,\dot \theta) \thinspace | \thinspace \dot \theta = 0 \thinspace or \thinspace \theta = (2p+1)\frac{\pi}{2}, p \in Z  \right\}
\end{align}
There are no trajectories in S that also lie in D apart from the trivial trajectory $E=0$, therefore by LaSalle's invariance principle, the swing up energy asymptotically converges to the desired energy level for all conditions in D.
An F defined in the following way makes $a_l$ equal to (24)
\begin{align}
    &F = (M+msin^2\theta)a_d+\nu_1 \nonumber \\
    &a_d = l \dot \psi^2 sin \theta cos \theta  - k_E \dot \theta cos \theta  E \nonumber \\
    &\nu_1 = mgsin\theta cos\theta + ml\dot \psi^2 sin\theta cos^2 \theta - ml \dot \theta^2 sin \theta  
\end{align}
We can now add a PD term to F defined above to regulate the cart to the origin. Since the nonholonomic constraints allow F to be only in the direction of orientation of the WMR, modifying F alone does not regulate the system to the origin.Therefore we design $\tau$ in such a way that the orientation of WMR is always aiming away from the origin. So the desired orientation at any instant can be defined as follows  
\begin{align}
tan(\psi_d) = y/x
\end{align}
We can then obtain the derivatives to be
\begin{align}
\dot \psi_d &= \frac{x \dot y - y \dot x}{x^2+y^2} \\
\ddot \psi_d &= \frac{(x^2+y^2)(x \ddot y - y \ddot x)-(x \dot y - y \dot x)(2x \dot x + 2y \dot y)}{(x^2+y^2)^2}
\end{align}

The errors in orientation and its derivatives can be defined as
\begin{align}
    e_{\psi} = \psi-\psi_d \\
    e_{\dot \psi} = \dot \psi - \dot \psi_d
\end{align}
We now define a lyapunov function to arrive at a controller that can asymptotically converge the orientation to the desired orientation
\begin{align}
    V_\psi = \frac{1}{2}k_{\psi}(\psi-\psi_d)^2+\frac{1}{2}k_{\dot \psi}(\dot \psi-\dot \psi_d)^2
\end{align}
for some positive $k_{\psi}$ and $k_{\dot \psi}$
\begin{align}
    \dot V_\psi = k_{\psi}(\dot \psi-\dot \psi_d)(\psi-\psi_d)+(\ddot \psi-\ddot \psi_d)(\dot \psi-\dot \psi_d)
\end{align}
To make $\dot V_{\psi}$ to be negative semi definite, we define
\begin{align}
    \tau = ml^2\dot \theta \dot \psi sin2\theta + (J+ml^2sin^2\theta)(-k_{\psi}e_{\psi}-k_{\dot \psi}e_{\dot \psi})
\end{align}
Define a set $S^1$ as follows
\begin{align}
    S^1 &= \left \{(\psi,\dot \psi) \thinspace | \thinspace \dot V_{\psi} = 0 \right\} \nonumber \\
    &= \left \{(\psi,\dot \psi) \thinspace | \dot \psi = \dot \psi_d \right\}
\end{align}
there are no trajectories in $S^1$ apart from the trivial trajectory $\psi = \psi_d$ and hence by LaSalle's invariance principle $\psi$ converges to $\psi_d$ asymptotically.
We now define the PD terms to be added to F so that the system is regulated to the origin. We restrict the PD gains to be in such a way that the dynamics is critically/over damped to prevent $\dot \psi_d$ from becoming infinity when the
WMR approaches the origin with nonzero velocity.
\begin{align}
F = (M+msin^2\theta)a_d+\nu_1 - k_v e_v - k_p e_p    
\end{align}
for some $k_v$ and $k_p$ satisfying the above mentioned conditions.

The velocity and position like error terms are defined as follows in a slightly unconventional way, the relevance of which will be understood in the next section.
\begin{align}
    e_v &= \dot x cos\psi + \dot y sin \psi \\
    e_p &= x cos \psi + y sin \psi
\end{align}
\section{Numerical simulations}
In this section we present numerical simulations of the system with initial conditions $[x,y,\psi,\dot x,\dot y,\dot \psi.\theta,\dot \theta] = [20,30,\pi,0.5,0,-1.5,\frac{\pi}{4},0]$, gains $[k_E,k_p,k_v,k_{\psi},k_{\dot \psi}]= [1,0.8,0.16,1,2]$ and system parameters $[M,m,J,l,g] = [1,0.1,0.01,1,9.81]$ 
\begin{figure}[h!]
	\centering
    \includegraphics[width=5in]{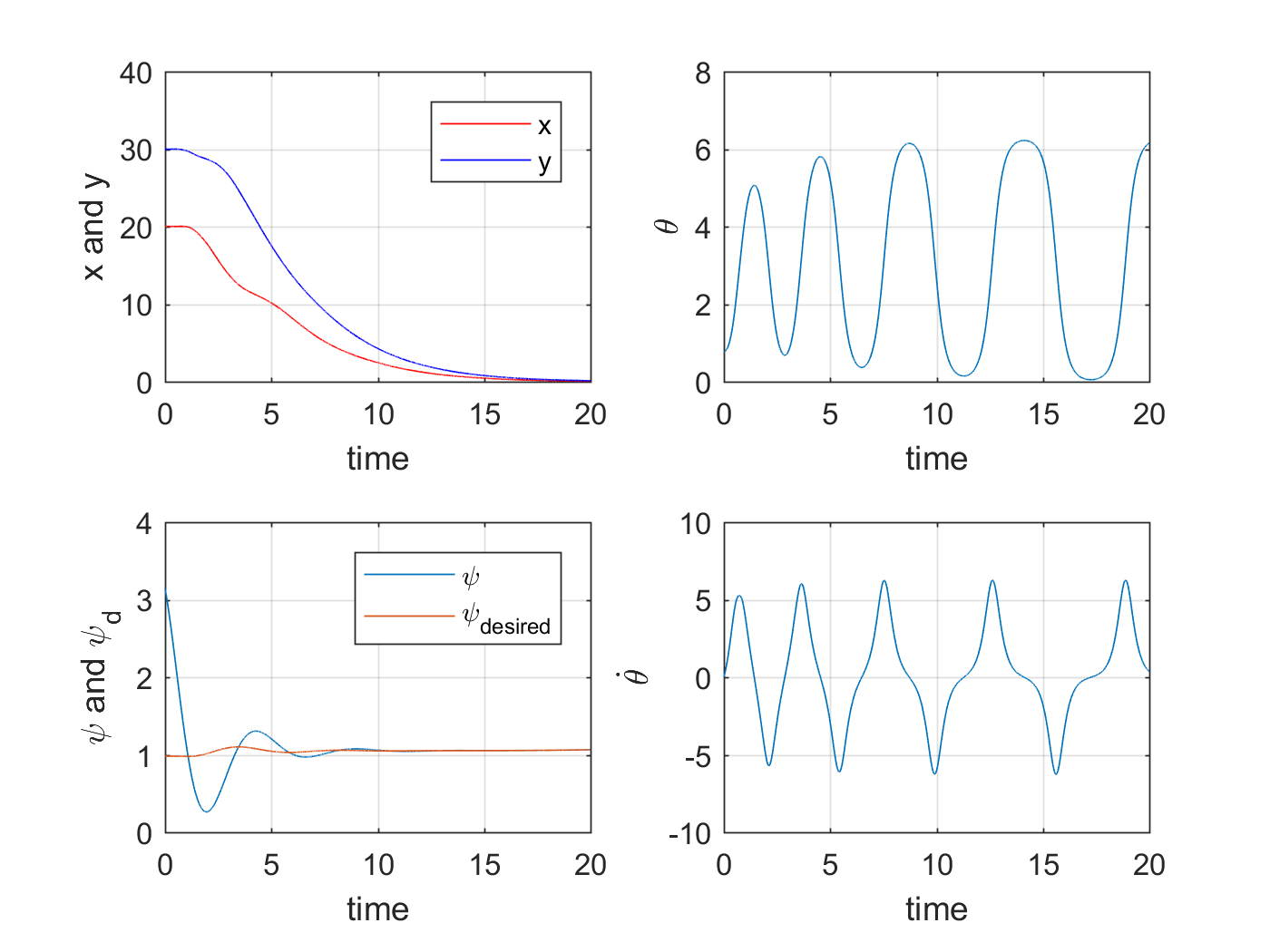}
\caption{\label{fig:Motion} Evolution of states with time}
	\label{schematic}
\end{figure}
\begin{figure}[h!]
	\centering
    \includegraphics[width=5in]{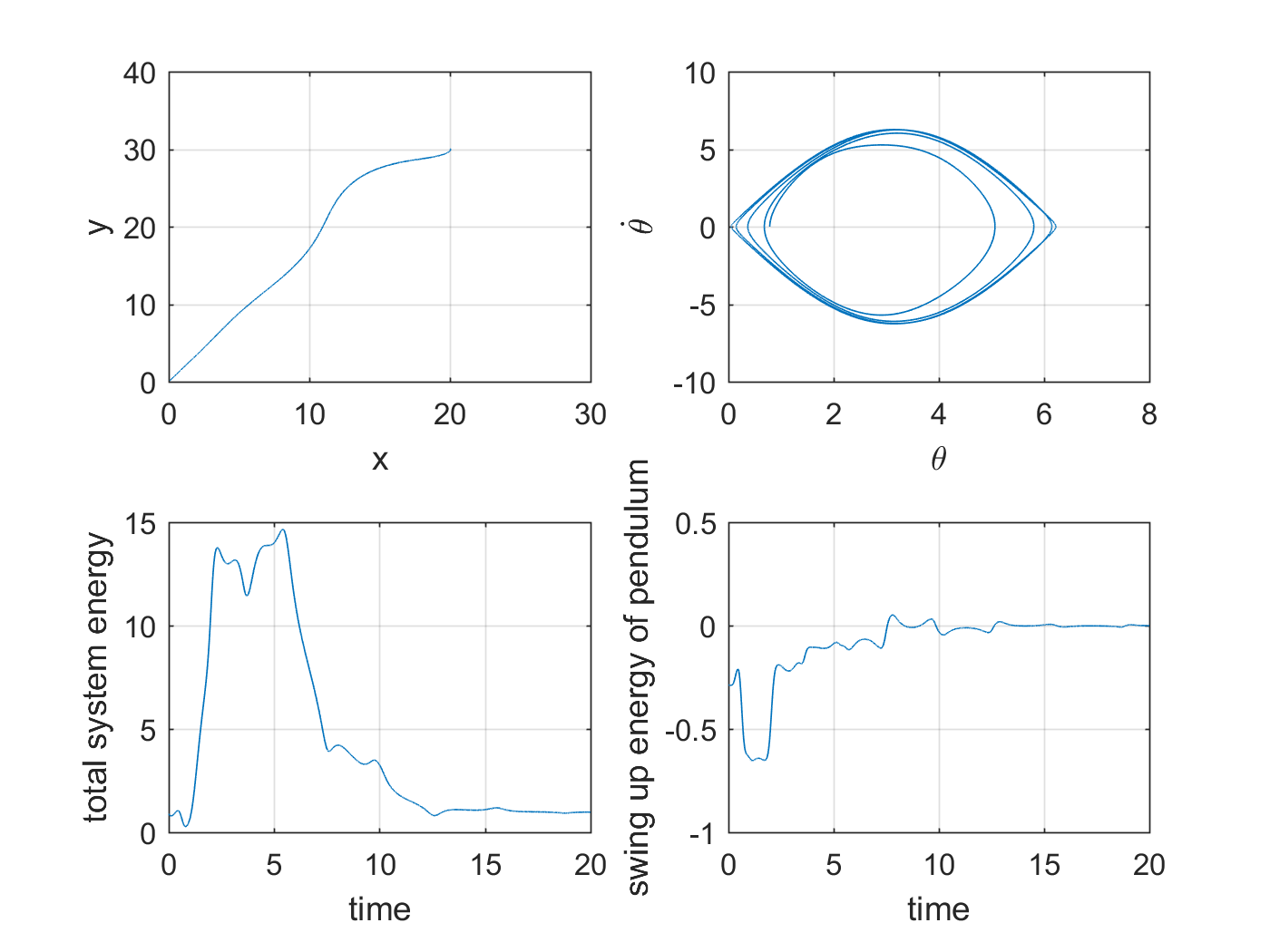}
\caption{\label{fig:Motion} Plots showing the path, phase portrait, total system energy versus time and the swing up energy versus time}
	\label{schematic}
\end{figure}

\section{Stability Analysis}
Since modifying F in (40) also modifies the way E evolves with time, it is not trivial to show that the desired task is achieved using Lyapunov analysis.
From the nonholonomic constraint that the WMR cannot move in the direction perpendicular to its orientation, we can solve for $\psi$ explicitly as
\begin{align}
 tan\psi = \frac{\dot y}{\dot x} 
\end{align}

In section 3, we have also shown that the actual orientation asymptotically converges to the desired orientation. Once the orientation converges
\begin{align}
tan\psi = tan\psi_d = \frac{y}{x}
\end{align}
Therefore after the convergence
\begin{align}
 \frac{y}{x} = \frac{\dot y}{\dot x} 
\end{align}
On rearranging, we obtain that 
\begin{align}
\dot x y - \dot y x = 0 
\end{align}
Substituting (46) in (32), we obtain that $\dot \psi_d$ and $\ddot \psi_d $ both converge to zero, implying that, $\psi_d$ and therefore $\psi$ tend to a constant value.
Assuming that at this constant value, $tan psi$ is finite, it can be seen from (44) that 
\begin{align}
    y = kx \\
    \dot y = k \dot x \\
    \ddot y = k \ddot x \\
    k = tan \psi 
\end{align}
Substituting (47) through (50) in the control law, at steady state we obtain
\begin{align}
    (\ddot x + k_v \dot x + k_p x) \sqrt{k^2+1} = - \frac{E \dot \theta cos \theta}{M+msin^2\theta}
\end{align}
which is an equation in one dimension and is analogous to the motion of cart pole. The proof of stability of this system has been presented using Floquet theory in [7] which the reader is encouraged to refer.


\bibliographystyle{alpha}
\bibliography{}
[1]Tang C, Miller PT, Krovi VN, Ryu J, Agrawal SK. Kinematic Control of Nonholonomic Wheeled Mobile Manipulator: A Differential Flatness Approach. ASME. Dynamic Systems and Control Conference, ASME 2008 Dynamic Systems and Control Conference, Parts A and B ():1117-1124. doi:10.1115/DSCC2008-2253. \newline
[2]Araki, N.; Okada, M.; Konishi, Y.; Ishigaki, and H.;. Parameter identification and
swing-up control of an acrobot system. International Conference on International
Technology, 2005.\newline
[3]Isabelle Fantoni and Rogelio Lozano. Non-linear Control for Underactuated Mechanical Systems. Communications and Control Engineering Series. Springer-Verlag,
2002.\newline
[4]K.J.Astrom, K.Furuta: Swinging up a pendulum by energy control,Automatica
Volume 36, Issue 2, February 2000, Pages 287-295\newline
[5]Arun D. Mahindrakar and Ravi N. Banavar. A swing-up of the acrobot based on a
simple pendulum strategy. International Journal of Control, 78(6):424–429, April
2005.\newline
[6]Mark Spong. The swingup control problem for the acrobot. IEEE Control Systems
Magazine, 15(1):49–55, February 1995.\newline
[7]Chung Choo Chung and John Hauser. Nonlinear control of a swinging pendulum.
Automatica, 31(6):851–862, June 1995.\newline
[8]I.Fantoni, R.Lozano, and Mark W. Spong: Energy Based Control of the Pendubot,IEEE Transactions on Automatic Control, VOL. 45, NO. 4, APRIL 2000, Pages 725-729. \newline
[9]Johan Åkesson, Karl Johan Åström, "Safe Manual Control of the Furuta Pendulum," Proceedings of the 2001 IEEE international Conference on Control Applications, pp.499-502, 2001\newline
[10]M. Iwase, K. J. Åström, K. Furuta and J. Åkesson "Analysis of safe manual control by using furuta pendulum," Proceedings of the 2006 IEEE International Conference on Control Applications, pp.568-572, 2006\newline
[11]Block D.J., Åström K.J., Spong M.W.: The reaction wheel pendulum. Synth. Lect. Control Mechatron. 1(1), 1–105 (2007)\newline
[12]R. M. Murray,M. Rathinam,W. sluis.Differential flatness of mechanical systems:A catalog of prototype systems.ASME \newline
[13]Bloch A.M. Nonholonomic Mechanics and Control, 2015 \newline
[14]M. van Nieuwstadt, M. Rathinam, R.M. Murray. Differential flatness and absolute equivalence \newline
[15]G. Walsh, D. Tilbury, S. Sastry, R. Murray,P. Laumond.Stabilization of systems with nonholonomic constraints. \newline
[16]J. Imura, K. Kobayashi, T. Yoshikawa, "Exponential stabilization problem of nonholonomic chained systems with specified transient response", Decision and Control 1996. Proceedings of the 35th IEEE Conference on, vol. 4, pp. 4733-4738 vol.4, 1996, ISSN 0191-2216.\newline
[17]A. M. Bloch, P. S. Krishnaprasad, J. E. Marsden, and R. M. Murray. Nonholonomic mechanical systems with symmetry.Technical Report CIT/CDS 94-013,California Institute of Technology,1994.
[18]L.Bushnell, D. Tilbury and S. Sastry.Steering Three-Input Chained Form Nonholonomic
Systems Using Sinusoids:The Fire Truck Example.Proceedings of the European Control Conference
Groningen, The Netherlands June 28 - July 1, 1993 \newline
[19]J.Hauser,S.Sastry,and G.Meyer.Nonlinear
control design for slightly nonminimum phase
systems|application to V/STOL aircraft.Automatica,
28(4):665{679, 1992.\newline
[20]I. Kanellakopoulos, P. V. Kokotovic, and A. S. Morse.
Systematic design of adaptive controllers for feedback
linearizable systems.IEEE Transactions on Automatic
Control, 36(11):1241{1253, 1991.\newline
[21]J-C. Latombe.Robot Motion Planning.Kluwer Academic
Publishers, Boston, 1991.\newline
[22]P.Martin.Contribution a l'Etude des Systems Differentiel
lement Plats.PhD thesis,L'Ecole Nationale
Superieure des Mines de Paris, 1992.\newline
[23]P. Martin.Endogenous feedbacks and equivalence.In
Mathematical Theory of Networks and Systems,Regensburg,
Germany, August 1993.\newline
[24]P. Martin.Personal communication, 1994.\newline
[25]P. Martin, S. Devasia, and B. Paden. A dierent look at
output tracking:Control of a VTOL aircraft.In Proc.
IEEE Control and Decision Conference,pages 2376{
2381, 1994.\newline
[26]P. Martin and P. Rouchon.Feedback linearization and
driftless systems.Mathematics of Control, Signals, and
Systems, 7(3):235{254, 1994.\newline
[27]M Fliess J Levine P Martin and P Rouchon On differentially flat nonlinear systems Comptes Rendus des Seances de lAcademie des Sciences "315:619-624 1992, Serie I. \newline
[28]M. Yamakita, M.Iwashiro, Y.Sugahara, K. Furuta, "Robust Swing Up Control of Double Pendulum," Proceedings of the American Control Conference, pp.290-295, vol.1, 1995

\end{document}